# OPTIC: **O**ptimizing **P**atient-Provider **T**riaging & **I**mproving **C**ommunications in Clinical Operations using GPT-4 Data Labeling and Model Distillation.

## AUTHORS

1. Alberto Santamaria-Pang, PhD, Microsoft Health AI, Redmond, WA, USA
Adjunct Faculty, Department of Biomedical Informatics & Data Science, Johns Hopkins University, Baltimore, MD, USA
2. Frank Tuan, Microsoft Health AI, Redmond, WA, USA
3. Ross Campbell, Health Information Technology, Johns Hopkins Medicine, Baltimore, MD, USA
4. Cindy Zhang, University of Washington, Seattle, WA, USA
5. Ankush Jindal MD, MS, Department of Biomedical Informatics & Data Science, Johns Hopkins University, Baltimore, MD, USA
6. Roopa Surapur, Microsoft, Baltimore, MD, USA
7. Brad Holloman, Microsoft, Baltimore, MD, USA
8. Deanna Hanisch, Health Information Technology, Johns Hopkins Medicine, Baltimore, MD, USA
9. Rae Buckley, PA-C, Department of Plastic and Reconstructive Surgery, Johns Hopkins Medicine, Baltimore, MD, USA
10. Carisa Cooney, MPH, Department of Plastic and Reconstructive Surgery, Johns Hopkins University, Baltimore, MD, USA
11. Ivan Tarapov, MS, Microsoft Health AI, Redmond, WA, USA
12. Kimberly S. Peairs, MD, Department of Medicine, Johns Hopkins University, Baltimore, MD, USA
13. Brian Hasselfeld, MD, Department of Medicine, Johns Hopkins University, Baltimore, MD, USA
14. Peter Greene, MD, Johns Hopkins Medicine, Baltimore, MD, USA

## ABSTRACT

**Background**: The COVID-19 pandemic has led to significant increases in the use of telemedicine as well as patient messaging via electronic medical portals (patient medical advice requests, or PMARs). While these systems offer benefits such as increased access to healthcare for patients, they have also created a substantial burden for healthcare providers due to surges in PMARs. This study addresses the need for an efficient tool for message triaging to reduce physician burden and improve patient-provider communication.

**Methods**: We developed OPTIC (Optimizing Patient-Provider Triaging & Improving Communications in Clinical Operations), a robust tool for message triaging using GPT-4 for data labeling and BERT for model distillation. The study utilized a dataset of 405,487 patient messaging encounters at Johns Hopkins Medicine between January to June 2020. We employed prompt engineering strategies with GPT-4 to create a high-quality labelled dataset, which was then used to train a BERT model for classifying messages as "Admin" or "Clinical".

**Results**: The BERT model achieved an accuracy of 88.85% on the test set derived from validated GPT-4 labeling, with a 88.29% sensitivity, 89.38% specificity, and an F1 score of 0.8842. BERTopic analysis identified 81 distinct topics within the test data, with the model achieving over 80% accuracy in classification for 58 topics. The system was successfully deployed through Epic's Nebula Cloud Platform, demonstrating its applicability in healthcare settings.

**Conclusion**: OPTIC offers a scalable and efficient tool for triaging PMARs in health systems. Leveraging advanced natural language processing techniques can reduce the administrative burden on healthcare providers and streamline patient care coordination.

# INTRODUCTION

The COVID-19 pandemic significantly impacted the global healthcare system and ways in which healthcare providers deliver care. To provide healthcare services safely and efficiently during the pandemic, telemedicine and patient portals, such as MyChart powered by Epic [1], became vitally important, and their use has continued post-pandemic. While MyChart can offer many benefits to patients (e.g., telemedicine appointments, prescription refills), it has also placed a substantial burden on providers. From December 2019 to December 2020, Holmgren et al found that patient messages to providers increased 157% [2]. During 2021, Primary Care Physicians at NYU Langone Health experienced a >30% increase in the volume of messages received per day [3]. Importantly, no natural mechanism currently exists to limit the InBasket messaging upsurge and these changes can lead to increased workloads, decreased efficiency in patient care management, and physician burnout [4-8]. As Verdean pointed out in a 2022 blog [9], MyChart messages can be viewed as the "Wild West of Patient Communication". Additionally, Schuetz [10] noted that responding to the most common patient messages, such as inquiries about test results or medication management, is time-consuming for care providers and some may not be adequately addressed electronically.

Given these insights, there is a need to design digital tools for PMAR categorization that reduce physician burden, optimize workflow, and improve patient-provider communication. Our main objective is to develop robust tools for patient portal systems like MyChart which facilitates message triaging by interpreting semantically meaningful clinical information. The scope for this analysis is to accurately classify between administrative and clinical messages that can potentially help in triaging (Figure 1).

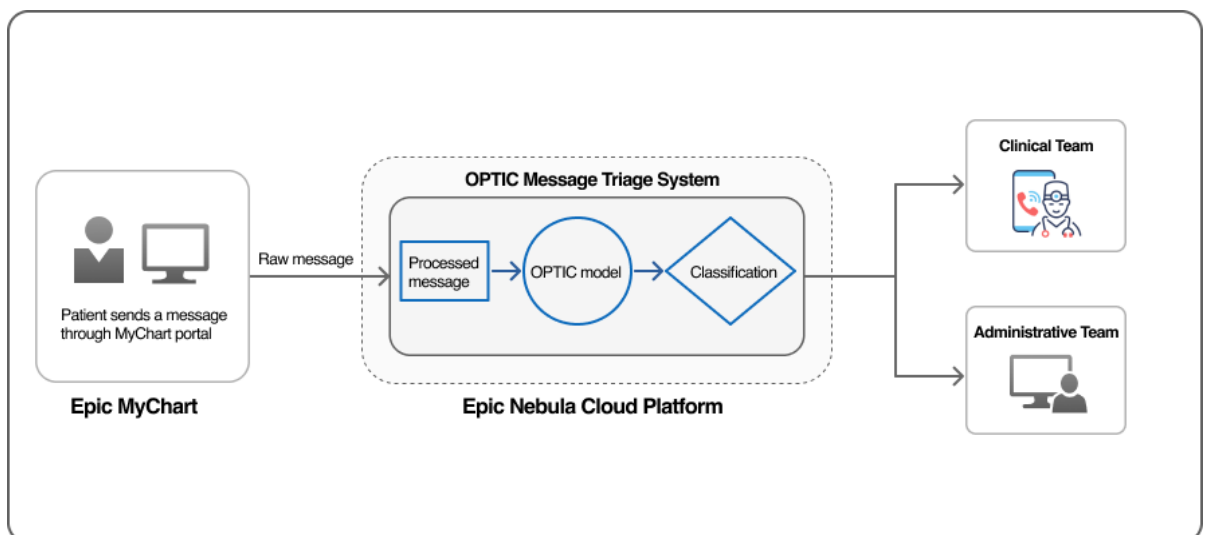

**Figure 1:** Overview of our Optimizing Patient-Provider Triaging & Improving Communications in Clinical Operations (OPTIC) system.

Telemedicine and asynchronous electronic care have become and will likely continue to be essential tools for providing healthcare service delivery. On average, physicians receive 40 InBasket messages per day through MyChart, with some receiving up to 150, leading to considerable time spent on

InBasket management [3]. While many of these messages are administrative in nature (such as appointment scheduling or routine medication refills), most healthcare messaging systems still lack the ability to prioritize messages based on clinical content or clinical acuity. This creates a significant challenge where highly clinical issues may be delayed in review while providers process high volumes of non-urgent administrative messages. Studies indicate that over 50% of these messages are administrative and could appropriately be handled by other healthcare team members [10, 11]. Without a structured system for sorting message types, urgent issues—whether clinical or administrative—can be delayed, underscoring the importance of more efficient communication systems.

Major challenges in implementing such a system include: 1) *Patient Variability*: Patients have unique backgrounds, medical histories, and communication styles, which can affect the way they communicate with their providers. **2)** *Complex Semantics*: Clinical language can be complex and domain-specific, which makes it challenging to accurately classify messages based on their content. The use of medical entities and acronyms further complicates understanding message meanings. In addition, many messages contain elements of both clinical needs and administrative requests, necessitating the creation of a definition and labeling hierarchy operationally relevant to real-world healthcare settings. **3)** *Interpretability*: BERT (Bidirectional Encoder Representations from Transformers) [12], can be difficult to interpret.

# LITERATURE REVIEW

For a review of medical emergency triage and patient prioritization in a telemedicine environment, please refer to [11]. A comparison of rule-based and machine learning (ML) approaches for classifying patient portal messages is presented in [13]. To the best of our knowledge, recent work of Large Language Models (LLMs) has primarily focused on medical dialog summarization [14-16].

On patients' message analysis, an ML model for classifying patient portal messages into four categories (straightforward, low, moderate, and no decision) was presented in [17]. The study used 500 message threads from primary care providers, comparing classifications with those of two subject matter experts (SMEs), and reported a model error rate of 36%. Sulieman et al., used word2vec to produce word embeddings and convolutional neural networks (CNNs) for classifying patient portal messages [18]. Similarly, in 2019 Wosik et al. presented models for identifying commonly asked questions by patients of their cardiology teams via MyChart [19]. In 2020, Judson et al. implemented patient self-triage and self-scheduling for COVID-19 [20], while Weber et al. (2022) reported an analysis on radiation oncology triage for nurses [21]. In 2021, De et al. [22] used topic modeling, employing Fast Health Care Interoperability Resources (FHIR)–Based Data Model to analyze patient message variability. In 2019, similar ML models were developed at the Mayo Clinic to classify patient messages into "Active Symptom" and "Logistic", based on over 4000 messages belonging to Dermatology, Cardiology and Gastro-enterology subspecialties, achieving an AUROC of 0.96 [23]. Regarding commercially available products for patient message triaging, TriageLogic [24] provides triage software designed for both nurses and call operators.

While previous methods and products address patient messages for optimizing clinical operations, they do not specifically address the automated triaging of clinical and administrative messages independently of the application (e.g., COVID-19 [20], radiation oncology [21], TriageLogic [24]). Furthermore, the applicability of LLMs seems to be limited to summarization tasks (e.g., [16]), and the reported accuracy of conventional ML models still requires improvement [17]. To address these limitations, we propose leveraging LLMs (BERT) to triage messages ("Admin" vs "Clinical") and further evaluate model performance and interpretability using topic discovery models (ex. BERTopic [25]).

This paper's unique contribution to advancing work in this area is the ongoing deployment of the model for testing by real outpatient providers (our SMEs) in collaboration with a health system using Epic. It documents the feasibility of building a model that performs with high confidence and accuracy in categorizing messages as "Admin" or "Clinical." The model is incorporated in a Nebula package, wherein Nebula Cloud Platform is Epic's Software as a Service (SaaS) cloud offering. The resulting package is deployed to Epic Hyperspace application client, the host for the MyChart patient portal system, making it potentially accessible to all systems using Epic. This model was developed not as an academic exercise but with the explicit goal of immediate implementation in the healthcare field.

The unique contributions that set our research apart are as follows:
1. LeveragingGPT-4 for data labeling through model distillation on a substantial dataset of over 400K messages) significantly larger than datasets used in prior studies.
2. The model, designed to confidently and accurately categorize messages as "Admin" or "Clinical", is actively deployed through Epic's Nebula Cloud Platform as part of their SaaS offering to ensure scalability. This underscores the practical applicability of the model in healthcare, moving beyond academic exercise to address real-world needs.

## DATA & COHORT SELECTION

IRB Review Statement: This project was acknowledged as *Not Human Subjects Research* by the Johns Hopkins Medicine Institutional Review Board (IRB00443265). Figure 2 provides an overview of the proposed methodology, detailing both the data selection and model development phases.

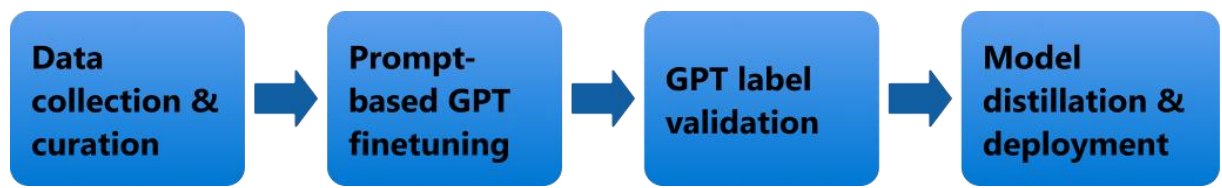

**Figure 2:** Overview of the proposed triaging system.

**Data Source**
An overview of data cleaning and processing is presented in Figure 3. The data was sourced from the MyChart/Epic electronic health record system at Johns Hopkins Medicine, spanning January to June 2020. It includes records from 405,487 patient messaging encounters in Primary Care Practice covering both Internal Medicine and Pediatric practices. The messages within Primary Care Practice encompass a broad range of activities: they serve as the initial contact point in the healthcare system and involve diagnosing, treating, and managing various health conditions. This practice area emphasizes preventive care, the management of chronic diseases, and the coordination of care with specialists. Moreover, it addresses both physical and mental health needs, ensuring consistent and ongoing care throughout a patient's life.
For Primary Care Practice, the initial message in each interaction is particularly important as it establishes the semantic context for the entire encounter. Therefore, we have chosen the first message from each interaction as its representative.

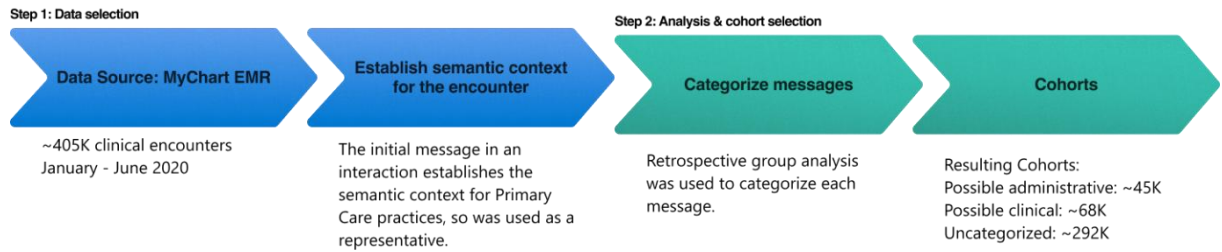

**Figure 3**: Data curation overview.

**Retrospective Group Analysis and Cohort Selection**
We conducted a retrospective analysis using the metadata from Epic and MyChart to categorize messages into three groups: "Possible Administrative," "Possible Clinical," and "Uncategorized." The **Administrative** group included 44,926 messages associated with employees (designated as 'EMP') who either lacked a service role (SER) or whose service role did not include a clinician title. The **Clinical** group consisted of 68,270 messages where healthcare providers contacted patients directly in conjunction with an Order or Note activity. The **Uncategorized** group, comprising 292,291 messages, included those that did not fit into either the Administrative or Clinical categories.

Examples of these messages include *"Good morning, I'm beginning to travel more for work, is the booster shot available at the Clinic?"* for "Admin" and *"Good Afternoon Dr., I had covid-19 and was vaccinated too. Do you suggest I get the boosters?"* for "Clinical." While the previous messages relate to COVID-19, they are representative of messages referring to different pathologies and diagnoses. The primary challenge lay in accurately interpreting message semantics that included both clinical and administrative elements. Figure 4 illustrates the clustering of all message embeddings into Administrative and Clinical groups, highlighting the complexity of distinguishing between these categories.

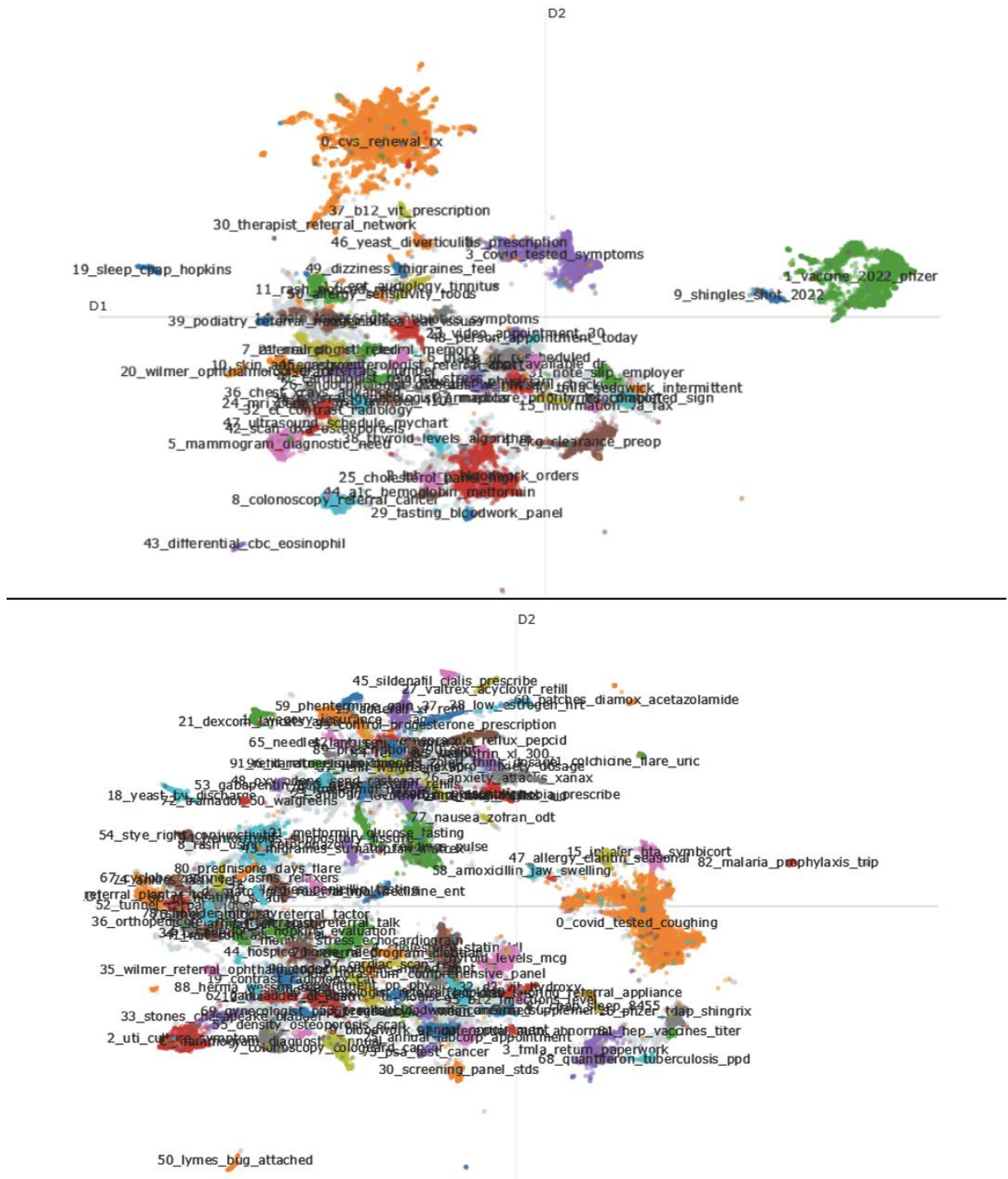

**Figure 4:** Clustering for Administrative(above) and Clinical(below) retrospective groups.

# GPT PROMPT ANALYSIS AND MODEL DISTILLATION

Our objective was to train a small language model (e.g., BERT) using GPT to establish high quality labeling dataset (~35K messages) generalizable to the full dataset (~405K messages). Our strategy involved the following steps: 1) GPT Few-shot prompting, 2) GPT prompt validation, and 3) model distillation (illustrated in Figure 5).

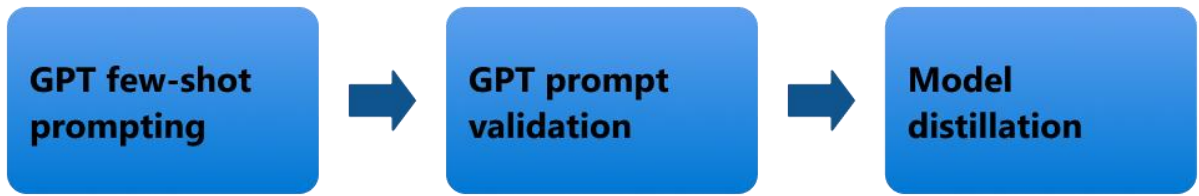

**Figure 5**: Prompt Engineering and Model Distillation Strategy.

1) **GPT Few-shot prompting**. We aim to sample representative Administrative and Clinical messages as evidence in a few-shot setting. We used the retrospective dataset (presented in the previous section) which comprises messages for Administrative and Clinical workflows. We used topic discovery to sample 200 messages distributed across clusters. A trained physician reviewed and validated messages used for the analysis. Examples of messages can be found in Table 1.

| Administrative Message Examples | Clinical Message Examples |
|---|---|
| • *"2nd COVID 19 Booster Shot. Hi Dr. Can you please update my medical record for my second COVID 19 booster shot."* Topic X.<br>• *"Insurance Form. Dr. Have you had time to look at the insurance form that completed for me? I need to turn it into the insurance company to be eligible for coverage."* Topic X.<br>• *"Jan appointment. I will have to reschedule this follow up appointment. Apologies for the late notice but woke up Tuesday with congestion and a headache that hasn't gone away. I've tested negative for covid so far but am not comfortable with an in-person visit. I had a full physical on Dec and have test results from the executive health program. These should have been sent to you as well."* Topic X. | • *"MRI Venogram. Hello, Thank for your review on the venogram. Can we do the 2nd option since it faster. Is there anything we need to do or is this something that you can request on your end? thanks!"* Topic X.<br>• *"Medication. Can you please give me something to relax my back and muscles in my body."* Topic X.<br>• *"Labs. Here are my labs. Is there any same day apts available tomorrow? I have been having a bad pain in my left side, it feels like there may be a knot there."* Topic X. |

Table 1. Example of Few-shot messages corresponding to Administrative and Clinical workflows.

We investigated four different strategies for prompt engineering. The strategies involved Zero-shot and Few-shot methods. In each scenario, the task was to classify a new message as "Administrative" or "Clinical" and explain the reason for such classification. Table 2 shows the Zero-shot prompt and general template of Few-shot prompt. The Zero-shot prompt prioritizes clinical and triaging, whereas the Few-shot prompts only provide examples of messages along with the corresponding category. In both cases, an explanation is required.

| | |
|---|---|
| Zero-shot | The following messages correspond to Admin or Clinical categories. For a new message, classify the message into Admin or Clinical.<br><br>If there are clinical symptoms listed that may require categorization to determine "urgent" or "not urgent" (where urgent can either be time urgency or acuity urgency), prioritize clinical.<br><br>Explain your reasoning. Provide output as: (Admin/Clinical), Explanation.) |
| Few-shot | The following messages correspond to Admin or Clinical categories:<br>    1. Message 1, Admin<br>    2. Message 2, Admin<br>    .<br>    .<br>  n. Message n, Admin<br><br>    1. Message 1, Clinical<br>    2. Message 2, Clinical<br>    .<br>    .<br>  n. Message n, Clinical<br>For a new message, classify the message into Admin or Clinical. Explain your reasoning. Provide output as: (Admin/Clinical), Explanation. |

Table 2. Prompt engineering strategies

Table 3 summarizes prompt engineering details, including prompt and GPT type along with number of representative messages selected. The number of messages were evenly distributed across Administrative and Clinical categories. Prompt experiment E1 and E3 utilized a random subset of the messages from prompt experiment E2.

| Prompt Experiment | Prompt Type | GPT Type | Number of messages |
|---|---|---|---|
| E1 | Few Shot | 4-32K | 10 |
| E2 | | 4-32K | 200 |
| E3 | | 3.5 Turbo | 120 |
| E4 | Zero Shot | 4-32K | NA |

Table 3. Prompt engineering design.

2) **GPT prompt validation.** To validate the different prompts, a validation dataset consisting of approximately 2,000 messages evenly distributed between Administrative and Clinical groups was established. The evaluation consisted of prompting the model with each message, then predicting a category (Administrative or Clinical) and providing an explanation as presented in Table 2. Three Physicians evaluated the prediction and provided explanation.

| Prompt Experiment | Accuracy | Sensitivity | Specificity | Precision | F1 score |
|---|---|---|---|---|---|
| E1 | 0.89 | 0.88 | 0.91 | 0.90 | 0.89 |
| **E2** | **0.99** | **0.99** | **1.0** | **1.0** | **0.99** |
| E3 | 0.85 | 0.75 | 0.96 | 0.95 | 0.84 |
| E4 | 0.65 | 0.32 | 0.98 | 0.95 | 0.48 |

Table 4. Performance metrics based on prediction of 2,000 messages along with provided explanations.

**3) Model distillation.** We conducted a series of four experiments (Table 4), with the best model corresponding to experiment E2 (GPT-4-32K with 200 input messages). Its accuracy was 0.99 when validated in 2,000 messages that were evenly distributed between the "Administrative" and "Clinical" groups. The scalability of such an approach faces hurdles, particularly when considering large-scale systems like Johns Hopkins Medicine which processes over a million in-basket messages across multiple specialties annually. To address cost constraints, we turned our attention to exploring models prioritizing computational efficiency. A small language model (e.g., BERT) emerged as a viable solution due to its ability to deliver robust performance on CPU, addressing economic considerations and environmental concerns related to carbon emissions. To harness the benefits of BERT, we retrained the model using messages labeled through the GPT approach.

To achieve this, we created a training dataset with GPT labels, sampling ~33K additional messages from our retrospective group analysis corresponding to the "Possible Administrative" and "Possible Clinical" groups. For the validation dataset, we used the same 2,000 messages which were used for GPT prompt validation. This retraining resulted in a noteworthy improvement in accuracy, reaching 92% thanks to the high quality of the labels obtained.

While the cost constraints associated with GPT-4 prompted the exploration of alternative models, particularly BERT, our findings showcase the promise of BERT in large-scale message classification systems. The BERT model not only demonstrated substantial accuracy improvements but also presented a dual-peak structure in the Kernel Density Estimation (KDE) plot, a statistical visualization that shows the probability density of the model's confidence scores. This bimodal distribution underscores the model's ability to clearly differentiate between different message types, contributing to a reliable inference mechanism.

# RESULTS

A prompt-based, fine-tuned version of GPT4-32K was used to label the dataset for model distillation. The whole dataset comprised of ~36k messages with equal proportions of "Admin" and "Clinical" messages. We split the dataset into three distinct datasets: a training set comprising 33,861 text samples, a validation set containing 3,387 text samples, and a test set with 3,454 text samples. Our experiments were conducted on AzureML using Python 3.7.3 and PyTorch 1.8.0, leveraging a single NVIDIA Tesla V100 GPU to accelerate the training process.

The model demonstrated an accuracy of 88.85%, a sensitivity of 88.29%, a specificity of 89.38%, and an F1 score of 0.8842. These metrics collectively indicate the model's strong ability to accurately identify and differentiate between the "Admin" and "Clinical" classes, showcasing its effectiveness in handling the nuances of medical communication classification.

To gain deeper insights into the model's performance across different types of content, BERTopic was employed to extract topics from the test set. This unsupervised topic modelling technique helped identify 81 distinct topics within the test data, providing a granular view of the message content. The model's accuracy was then assessed for each of these topics individually. Out of the 81 identified topics, the model achieved over 80% accuracy for 58 topics [Appendix Figure 8]. The model maintains high performance across a diverse range of subject matters which suggests that it has developed a robust understanding of various medical and administrative concepts, allowing it to generalize well beyond just a few dominant themes in the dataset.

# CONCLUSION

This paper addresses the critical need for efficient triaging of patient medical advice requests within patient portal systems like MyChart, particularly given the increasing and persistent demands for virtual healthcare following the COVID-19 pandemic. Leveraging a dataset of 405,487 encounters from Johns Hopkins Medicine spanning Primary Care Practice from January to June 2020, we developed and deployed a BERT-based model for classifying messages as "Admin" or "Clinical" through Epic's Nebula Cloud Platform. This model not only demonstrates high accuracy in categorization but also underscores its practical deployment potential within the healthcare sector.

Our main contributions include the development of a scalable triaging tool integrated into Epic's infrastructure, enhancing operational efficiency and provider workflow management. This system represents an improvement over the historical baseline state where no automated assessment of message content existed, effectively reducing the burden on healthcare providers while also enhancing patient care coordination by streamlining message sorting. Deployment through Nebula as part of Epic's SaaS offering ensures accessibility and scalability across healthcare systems. This research underscores the potential of pairing Large Language Models like GPT-4 for large scale data labelling and with small language models like BERT for deployment on edge devices ensuring cost-effectiveness for systems similar in scale to Johns Hopkins Medicine.

Future work will focus on enhancing interpretability and expanding the model's capabilities to handle additional nuances in patient communication, particularly as this may apply to other clinical departments across Johns Hopkins Medicine. Ultimately, our research highlights the transformative potential of leveraging advanced natural language processing techniques to address healthcare challenges and improve patient outcomes.

# ACKNOWLEDGEMENTS

We would like to express our sincere gratitude to Alexandra Eikenbary for her contributions in creating the figures for this manuscript. We also extend our thanks to the Johns Hopkins Health IT team for their ongoing support throughout the duration of this project.

# Appendix

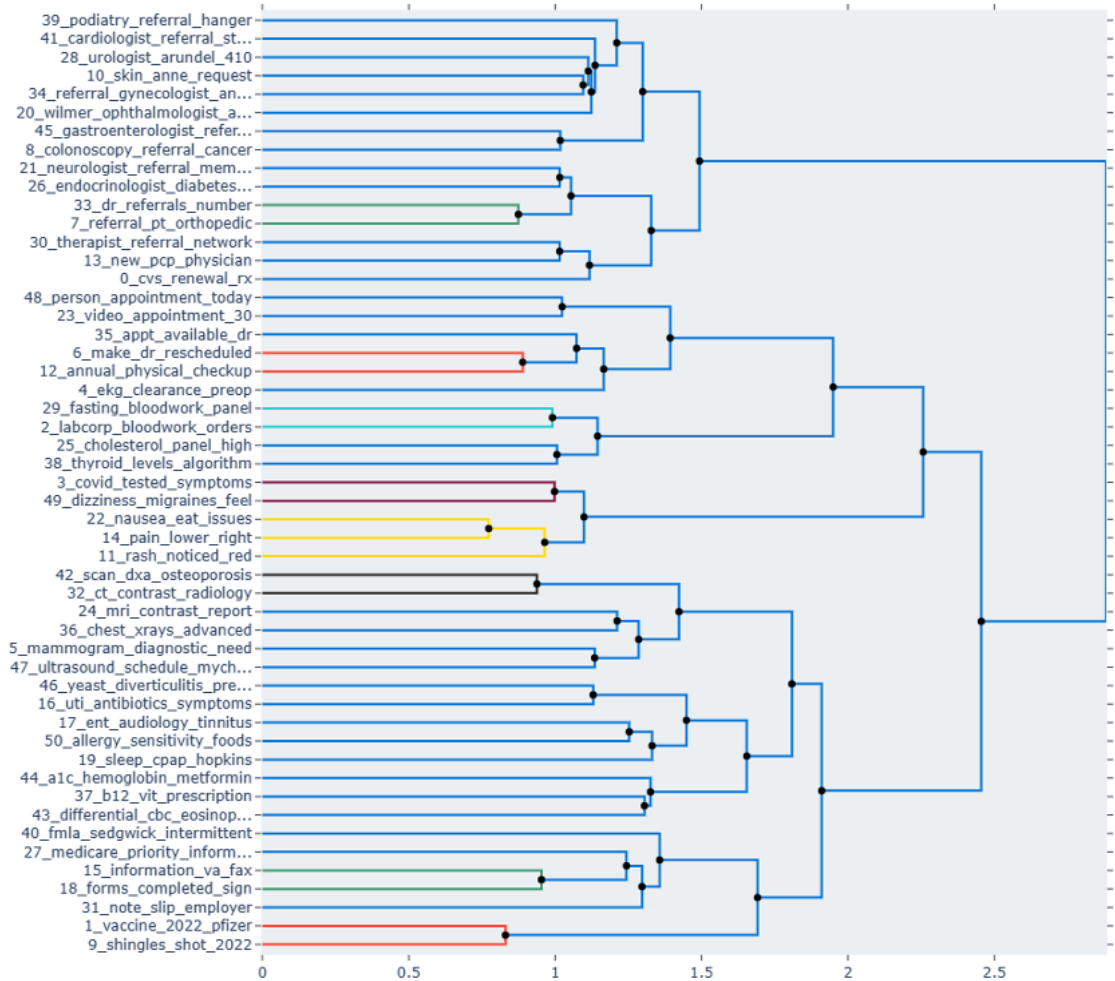

**Figure 6:** Hierarchical clustering for Admin retrospective group.

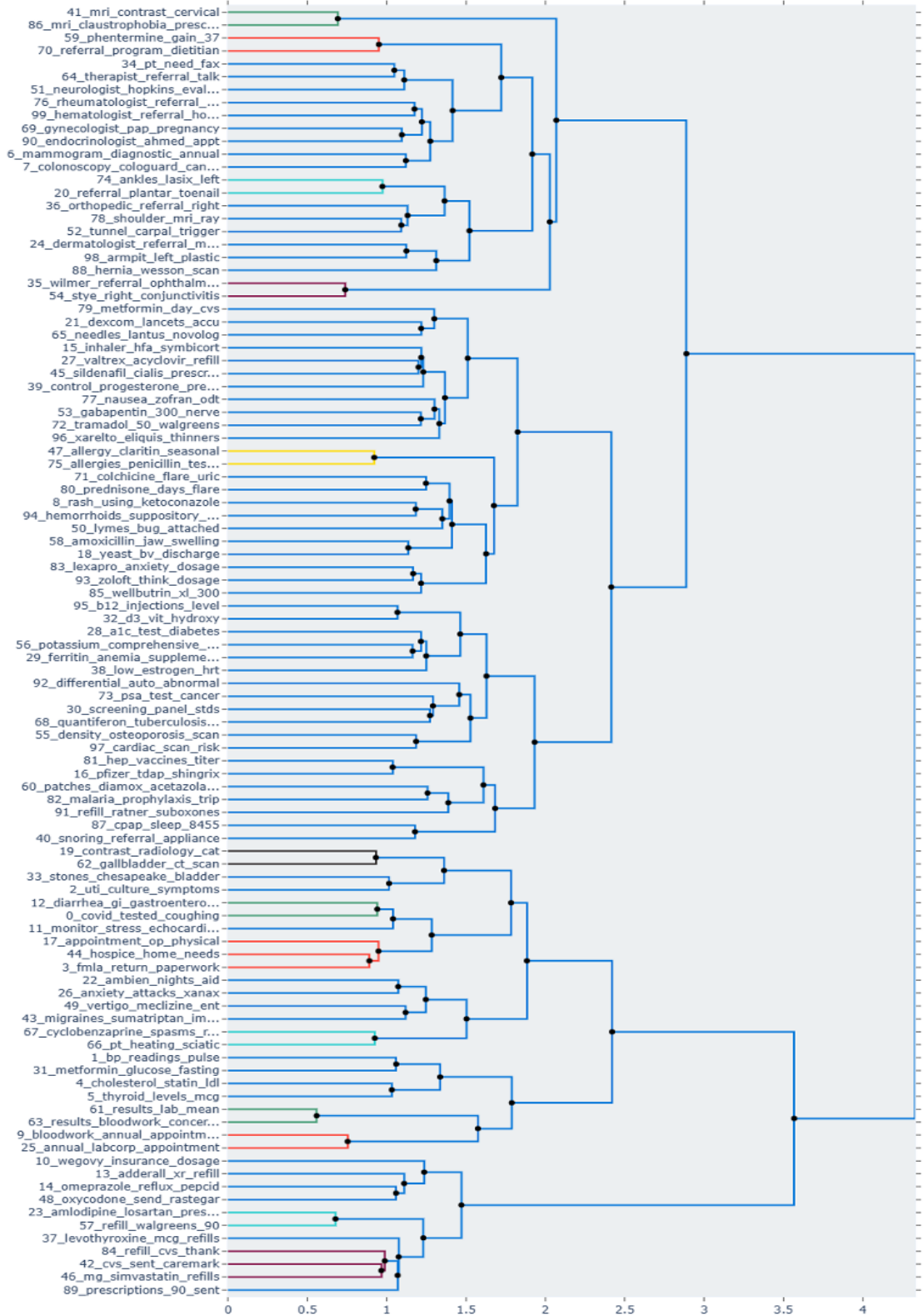

**Figure 7**: Hierarchical clustering for Clinical retrospective group.

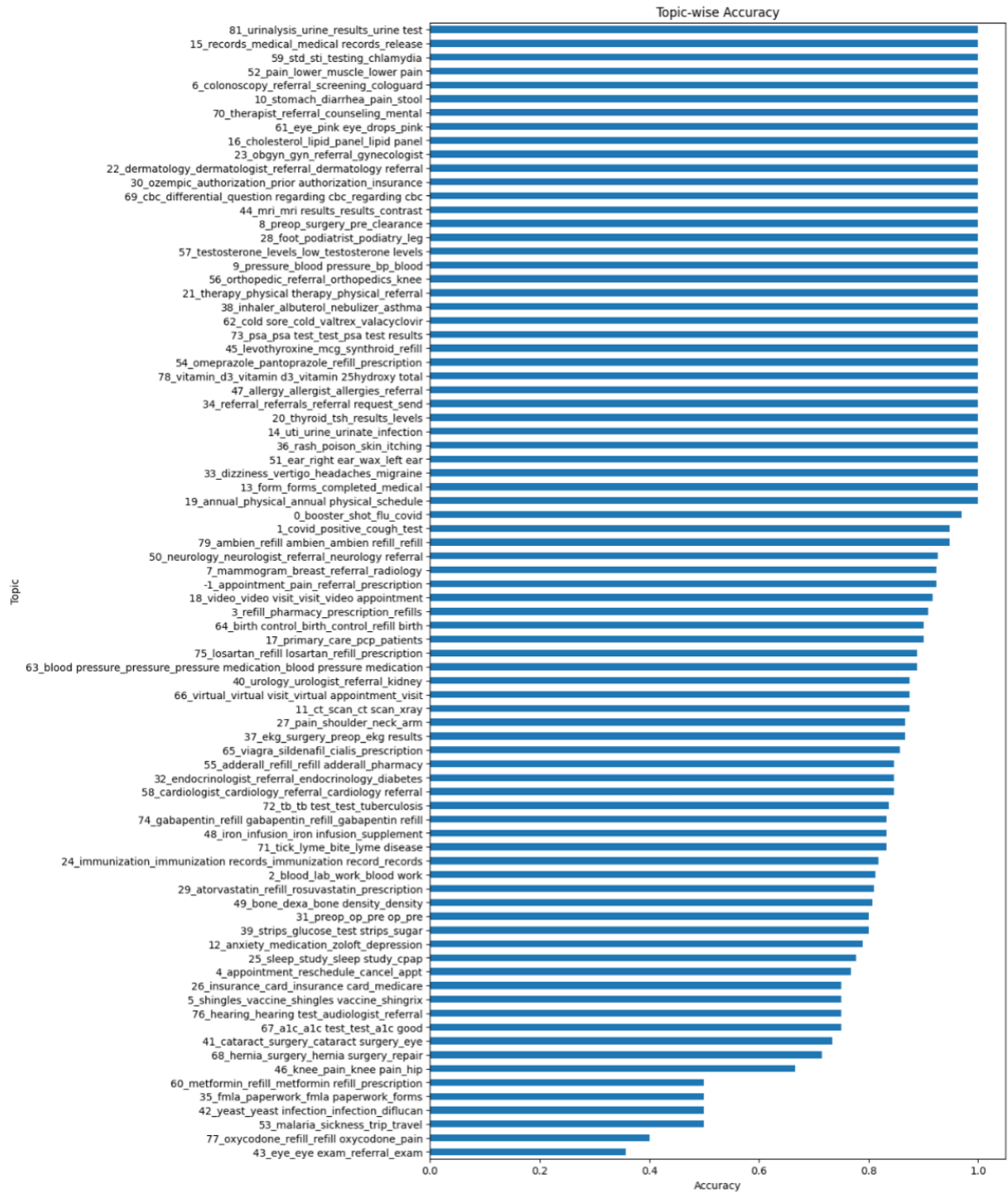

**Figure 8**: Topic wise accuracy of BERT model on test set.

# REFERENCES

[1]    https://mychart.org/

[2]    Holmgren A, Downing N, Tang M, et al. Assessing the impact of the COVID-19 pandemic on clinician ambulatory electronic health record use. *J Am Med Inform Assoc*. 2022;29(3):453-460. doi: 10.1093/jamia/ocab268.

[3]    N. L. H. T. Hub, "How We're Improving Physicians' Messaging Experience Through Digital Tools," NYU Langone Health Tech Hub. Accessed: Nov. 15, 2023. [Online]. Available: https://medium.com/nyu-langones-health-tech-hub/how-were-improving-physicians-messaging-experience-through-digital-tools-1c0abd8e711b

[4]    Sieck C, Hefner J, Schnierle J, et al. The Rules of Engagement: Perspectives on Secure Messaging From Experienced Ambulatory Patient Portal Users. *JMIR Med Inform*, vol. 5, no. 3, p. e13, Jul. 2017, doi: 10.2196/medinform.7516.

[5]    Ramsey A, Lanzo E, Huston-Paterson H, et al. Increasing Patient Portal Usage: Preliminary Outcomes from the MyChart Genius Project. *J Adolesc Health*, vol. 62, no. 1, pp. 29–35, Jan. 2018, doi: 10.1016/j.jadohealth.2017.08.029.

[6]    Kelley L, Phung M, Stamenova V, et al. Exploring how virtual primary care visits affect patient burden of treatment. *International Journal of Medical Informatics*, vol. 141, p. 104228, Sep. 2020, doi: 10.1016/j.ijmedinf.2020.104228.

[7]    West C, Dyrbye L, and Shanafelt T. Physician burnout: contributors, consequences and solutions. *Journal of Internal Medicine*, vol. 283, no. 6, pp. 516–529, 2018, doi: 10.1111/joim.12752.

[8]    Hilliard RW, Haskell J, Gardner RL. Are specific elements of electronic health record use associated with clinician burnout more than others? *J Am Med Inform Assoc*. 2020 Jul 1;27(9):1401-1410. doi: 10.1093/jamia/ocaa092. PMID: 32719859; PMCID: PMC7647296.

[9]    "MyChart Messages the Wild West of Patient Communication," 33 Charts. Accessed: Mar. 04, 2022. [Online]. Available: https://33charts.com/mychart-messages-the-wild-west-of-patient-communication/

[10]   Schuetz S. Impact of MyChart Communication on Provider Burden. *Family Medicine Clerkship Student Projects*, Jan. 2021, [Online]. Available: https://scholarworks.uvm.edu/fmclerk/693

[11]   Napi NM, Zaidan AA, Zaidan BB, et al. Medical emergency triage and patient prioritisation in a telemedicine environment: a systematic review. *Health Technol.*, vol. 9, no. 5, pp. 679–700, Nov. 2019, doi: 10.1007/s12553-019-00357-w.

[12]    Devlin J, Chang MW, Lee K. et al. Bert: Pre-training of deep bidirectional transformers for language understanding. *arXiv preprint arXiv:1810.04805* (2018).

[13]   Cronin RM, Fabbri D, Denny JC, et al. A comparison of rule-based and machine learning approaches for classifying patient portal messages. *International Journal of Medical Informatics*, vol. 105, pp. 110–120, Sep. 2017, doi: 10.1016/j.ijmedinf.2017.06.004.

[14]   Molenaar S, Maas L, Burriel V, et al. Medical Dialogue Summarization for Automated Reporting in Healthcare. *Advanced Information Systems Engineering Workshops*, S. Dupuy-Chessa and H. A. Proper, Eds., in Lecture Notes in Business Information Processing. Cham: Springer International Publishing, 2020, pp. 76–88. doi: 10.1007/978-3-030-49165-9_7.

[15]   Joshi A, Katariya N, Amatriain X, et al. Dr. Summarize: Global Summarization of Medical Dialogue by Exploiting Local Structures. *arXiv:2009.08666 [cs]*, Sep. 2020, Accessed: Mar. 05, 2022. [Online]. Available: http://arxiv.org/abs/2009.08666

[16]   Chintagunta B, Katariya N, Amatriain X, et al. Medically Aware GPT-3 as a Data Generator for Medical Dialogue Summarization. *Proceedings of the 6th Machine Learning for Healthcare Conference*, PMLR, Oct. 2021, pp. 354–372. Accessed: Mar. 05, 2022. [Online]. Available: https://proceedings.mlr.press/v149/chintagunta21a.html

[17]   Patient Engagement HIT, "More Data Key to Understand AI for Sorting Patient Portal Messages,". Accessed: Mar. 07, 2023. [Online]. Available: https://patientengagementhit.com/news/more-data-key-to-understand-ai-for-sorting-patient-portal-messages

[18] Sulieman L, Glimore D, French C, et al. Classifying patient portal messages using Convolutional Neural Networks. *Journal of Biomedical Informatics*, vol. 74, pp. 59–70, Oct. 2017, doi: 10.1016/j.jbi.2017.08.014.

[19] Wosik J, Si S, Henao R, et al. Artificial Intelligence to Identify Commonly Asked Questions via an Electronic Patient Portal - Lessons From a Cardiology Department Within a Large Health System [abstract]. *Circulation*, vol. 140, no. Suppl_1, pp. A14805–A14805, Nov. 2019, doi: 10.1161/circ.140.suppl_1.14805.

[20] Judson TJ, Odisho AY, Neinstein AB, et al. Rapid design and implementation of an integrated patient self-triage and self-scheduling tool for COVID-19. *Journal of the American Medical Informatics Association*, vol. 27, no. 6, pp. 860–866, Jun. 2020, doi: 10.1093/jamia/ocaa051.

[21] Weber BW, Blitzer GC, Anderson BM, et al. Analysis of Patient Contacts with the Radiation Oncology Triage Nurse: The Experience of a Single Center. *International Journal of Radiation Oncology, Biology, Physics*, vol. 114, no. 1, p. e25, Sep. 2022, doi: 10.1016/j.ijrobp.2022.06.087.

[22] De A, Huang M, Feng T, et al. Analyzing Patient Secure Messages Using a Fast Health Care Interoperability Resources (FIHR)–Based Data Model: Development and Topic Modeling Study. *Journal of Medical Internet Research*, vol. 23, no. 7, p. e26770, Jul. 2021, doi: 10.2196/26770.

[23] Ahmed PT, Sunyang F, Aditya K, et al. Artificial intelligence to organize patient portal messages: a journey from an ensemble deep learning text classification to rule-based named entity recognition. IEEE International Conference on Bioinformatics and Biomedicine (BIBM), pp. 1380-1387, Nov 2021, doi: 10.1109/BIBM47256.2019.8982942.

[24] "Medical Call Center Software, Nurse Triage on Call," TriageLogic: Remote Nurse Triage Software and Services. Accessed: Mar. 07, 2023. [Online]. Available: https://triagelogic.com/

[25] Grootendorst M. BERTopic: Neural topic modeling with a class-based TF-IDF procedure. *arXiv*, Mar. 11, 2022. doi: 10.48550/arXiv.2203.05794.